\documentclass[10pt,twocolumn,letterpaper]{article}
\usepackage[pagenumbers]{cvpr}
\makeatletter
\renewcommand\LenToUnit[1]{\strip@pt\dimexpr#1\relax}
\makeatother
\usepackage{graphicx}
\usepackage{adjustbox}
\usepackage{xspace}
\usepackage{amsmath,amssymb}
\usepackage{booktabs}
\usepackage{multirow}
\usepackage{array}
\usepackage{xcolor}
\usepackage{microtype}
\usepackage{url}
\usepackage[pagebackref,breaklinks,colorlinks,allcolors=blue]{hyperref}
\usepackage[capitalize,nameinlink]{cleveref}

\newcommand{\method}{Gap2Co\xspace}

\def\confName{CVPR}
\def\confYear{2027}
\title{Gap2Co: Learning Residual Collaboration Demand \\
for Temporal 3D Semantic Occupancy Prediction}
\author{Zhiguo Zhao \quad Peitong Wu \quad Siyi Yan \quad Xiaoxi Hu \quad Kun Chen\\
Cong Zhao \quad Cairong Zhao \quad Yuchuan Du \quad Yuxiong Ji}
\begin{document}
\maketitle
\begin{abstract}
Collaborative 3D semantic occupancy prediction benefits from
complementary viewpoints but faces several key limitations.
Existing methods often underutilize historical ego and
neighbor observations for temporal scene completion,
while conventional communication criteria do not explicitly
capture the remaining benefit of new evidence after
temporal completion. Moreover, evidence acquisition and fusion
lack unified demand guidance.
To address these challenges, we propose \method, a
history-conditioned, demand-driven collaborative occupancy
framework.
First, temporal completion integrates historical ego features
and previously received neighbor evidence to recover missing
scene information.
Residual Demand Learning (RDL) then estimates what
collaboration is still needed, supervised by regional
occupancy-loss improvements from a full-information teacher.
Finally, the Residual Demand-Guided Landmarks (RDG-Landmarks) module
uses the predicted demand to prioritize byte-budgeted
requests and organize received evidence into shared summaries
and fine-grained supports.
Shared-support Joint Attention (SJA) enables query-specific
fusion of the organized evidence.
Experiments on SemanticOPV2V demonstrate that \method
achieves 45.74\% mean intersection over union (mIoU) and 80.95\% occupancy IoU,
outperforming VQSOP by 4.56 mIoU points while averaging
54.24~KiB/frame.
Extensive ablations and robustness evaluations further
validate the effectiveness of the proposed framework.
\end{abstract}

\section{Introduction}
\label{sec:introduction}

\begin{figure}[t]
    \centering
    \includegraphics[width=\columnwidth,clip]{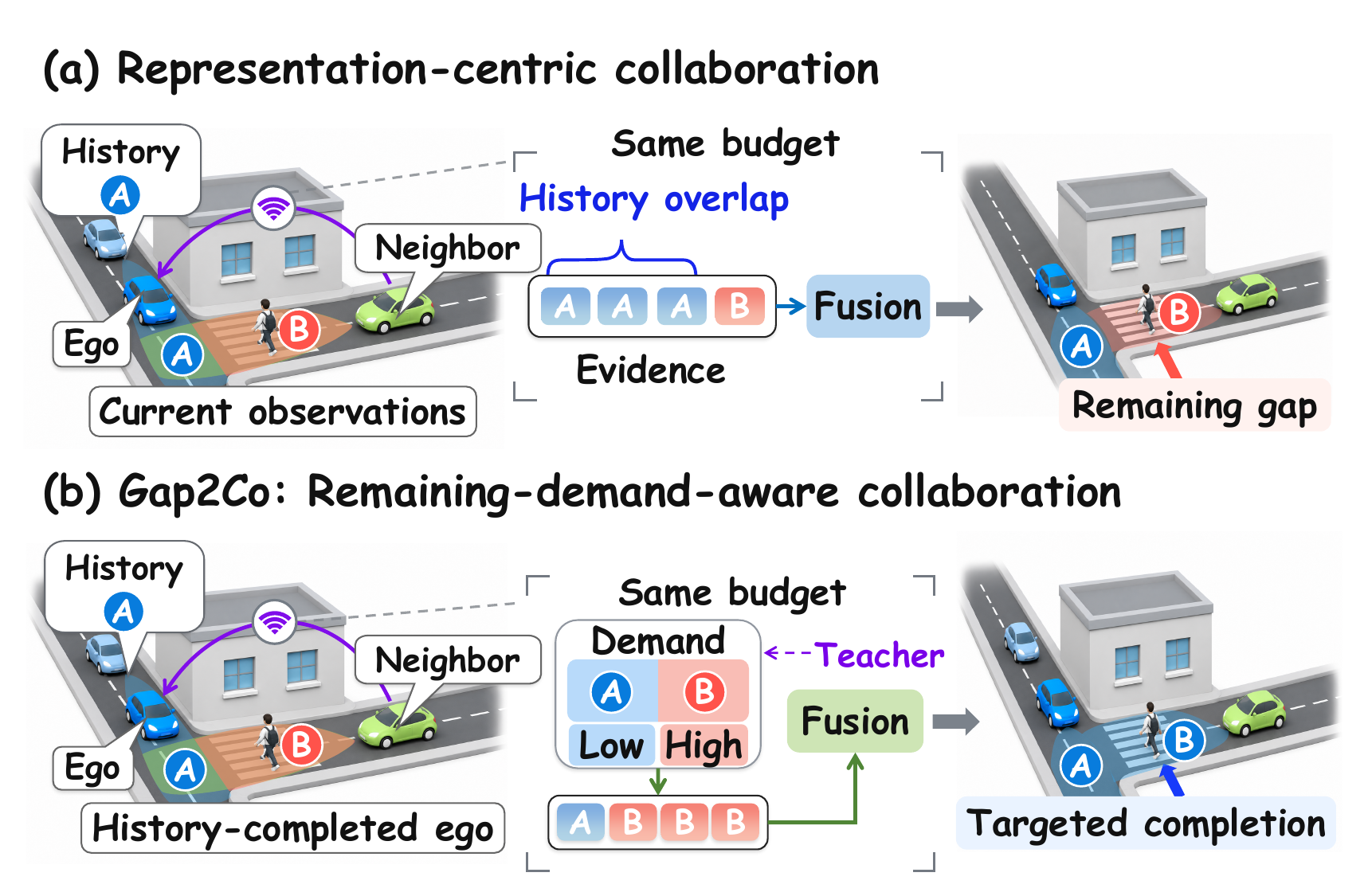}
    \caption{
    Motivation for remaining-demand-aware collaboration.
    (a) Compact representations~\cite{song2024cohff,chen2026vogs,li2026vqsop}
    may repeat evidence already recovered from history (A),
    leaving a gap (B).
    (b) After temporal completion using ego and neighbor histories,
    \method predicts the remaining demand to guide evidence
    acquisition and fusion under the same budget.
    Teacher supervision is used only during training.
    }
    \label{fig:motivation}
\end{figure}

Multi-agent collaborative perception extends a vehicle's observations
beyond its own viewpoint, improving 3D semantic occupancy prediction
under occlusion, limited fields of view, and weak distant observations.
However, exchanging high-dimensional 3D features incurs substantial
communication and computation costs~\cite{xu2023cobevt,wang2023core,li2026vqsop}.
From an entropy-reduction perspective, the key is not merely to
transmit compact representations, but to acquire complementary
evidence that provides task-relevant information beyond what
the receiver has already recovered~\cite{hu2024codefilling}.

Existing collaborative occupancy methods emphasize efficient
representation and communication.
CoHFF exchanges compressed orthogonal-plane features~\cite{song2024cohff},
GSFusion uses sparse semantic Gaussians~\cite{chen2026vogs},
and VQSOP combines sparse-aware vector quantization with adaptive
spatial refinement~\cite{li2026vqsop}.
Despite these advances, three challenges remain.
\textbf{First, temporal information remains underutilized.}
While temporal modeling has been explored in single-agent occupancy
and general collaborative perception~\cite{li2024viewformer,tang2025cost,wang2024panoocc,leng2025stocc,hong2024mrcnet},
collaborative 3D occupancy methods largely focus on current cross-agent
observations, without explicitly combining historical ego features
and previously received neighbor evidence for scene completion.
As a result, newly transmitted information may overlap with
what the receiver has already recovered from history,
limiting the additional information gain
(Fig.~\ref{fig:motivation}).

\textbf{Second, remaining collaboration demand is not explicitly modeled.}
Existing communication strategies prioritize information using
spatial importance~\cite{hu2022where2comm},
task relevance~\cite{hu2026pragcomm}, or temporal
redundancy~\cite{tang2025cost}.
However, these criteria do not directly estimate how much new
neighbor evidence can still improve occupancy after temporal
completion, when accumulated observations have already recovered
part of the scene information.
\textbf{Third, evidence acquisition and fusion lack unified guidance.}
Selecting useful information alone does not ensure its effective
utilization~\cite{wang2023umc}.
Although structured attention mechanisms organize features
through landmarks and reusable supports~\cite{xiong2021nystromformer,wen2026mita},
their organization is not explicitly conditioned on the receiver's
remaining collaboration demand.
These challenges motivate a unified framework that determines
\emph{what has already been recovered}, \emph{what is still needed},
and \emph{what evidence to acquire and how to use it}.

To address these challenges, we propose \method, a
history-conditioned, demand-driven framework for collaborative
3D semantic occupancy prediction.
To exploit accumulated observations and avoid redundant collaboration,
we first complete the ego representation using pose-aligned
historical ego features and previously received neighbor evidence,
establishing \emph{what the receiver has already recovered}.
To identify \emph{what is still needed}, we introduce
\textbf{Residual Demand Learning (RDL)}.
A frozen full-information teacher compares paired occupancy
predictions with and without new collaborative evidence under
the same completed ego state.
Motivated by further entropy reduction, we use the resulting
regional occupancy-loss reduction as a task-grounded signal
to supervise remaining collaboration demand.
RDL thus estimates the potential benefit of additional evidence
without teacher access at inference.

To translate the predicted demand into effective collaboration,
we introduce \textbf{Residual Demand-Guided Landmarks
(RDG-Landmarks)}.
The predicted demand prioritizes regional requests under a
serialized-byte budget and guides the construction of
demand-weighted 3D landmarks.
These landmarks organize received neighbor evidence into
shared summaries and landmark-specific fine supports.
Building on Mixture-of-Top-$k$ Attention (MiTA)~\cite{wen2026mita},
\textbf{Shared-support Joint Attention (SJA)} enables multiple
voxel queries to share supports while independently attending
to summarized context and fine-grained evidence.
In this way, the same demand signal guides both
\emph{what evidence to acquire} and \emph{how to use it}.

On SemanticOPV2V, our framework achieves \textbf{45.74\% mIoU} and
\textbf{80.95\% occupancy IoU}, outperforming VQSOP
by \textbf{4.56 mIoU points} at an average communication
cost of \textbf{54.24~KiB/frame}.
Under a 32~KiB/frame communication cap, it already reaches
45.72\% mIoU.
Ablation and robustness studies assess
temporal completion, residual demand learning,
evidence organization, and historical reuse.

Our contributions are summarized as follows:
\begin{itemize}
    \item To the best of our knowledge, we are the first to
    jointly leverage historical ego features and previously
    received neighbor evidence for temporal completion
    in collaborative 3D semantic occupancy prediction.
    Building on this completed representation, we introduce
    Residual Demand Learning (RDL), which learns the remaining
    task benefit of new collaborative evidence through
    teacher-supervised regional occupancy-loss reduction.

    \item We develop a unified demand-driven acquisition and fusion
    framework. RDG-Landmarks uses predicted demand to prioritize
    byte-budgeted requests and organize received evidence
    into shared summaries and fine supports, while SJA
    enables query-specific evidence fusion.

    \item Extensive experiments on SemanticOPV2V demonstrate
    improved occupancy accuracy and a favorable
    accuracy--communication trade-off, with ablation and
    robustness studies validating the proposed components
    under communication constraints, pose errors,
    delayed observations, and occlusion.
\end{itemize}

\section{Related Work}
\label{sec:related_work}

\subsection{3D Semantic Occupancy Prediction}
3D semantic occupancy prediction jointly estimates scene geometry and
semantic labels on a voxel grid. Camera-based methods organize image
evidence through tri-perspective planes~\cite{huang2023tpv}, sparse-to-dense
voxel prediction~\cite{li2023voxformer}, or sparse semantic
Gaussians in GaussianFormer (GSFormer)~\cite{huang2024gaussianformer}. Temporal modeling further
integrates historical observations~\cite{li2022bevformer,li2024viewformer}.
These representations support scene reconstruction from a single agent's
observations. Extending occupancy prediction across vehicles also requires
compact messages and selective use of complementary evidence to control
the cost of exchanging 3D features.

\subsection{Collaborative Perception}
Collaborative perception integrates information across agents through
raw-data, prediction, or intermediate-feature fusion.
V2X-ViT employs multi-agent attention for heterogeneous and asynchronous
observations~\cite{xu2022v2x}, while Where2comm selects informative
regions using spatial confidence~\cite{hu2022where2comm}.
PragComm explores spatiotemporal message selection and receiver-driven
communication~\cite{hu2026pragcomm}, and CoST exploits historical
information to reduce redundant transmissions~\cite{tang2025cost}.
For collaborative 3D semantic occupancy, CoHFF exchanges compressed
plane features~\cite{song2024cohff}, GSFusion fuses semantic
Gaussians~\cite{chen2026vogs}, and VQSOP employs sparse-aware vector
quantization with adaptive spatial refinement~\cite{li2026vqsop}.
Unlike these predominantly current-frame collaborative occupancy methods,
Gap2Co jointly exploits historical ego and neighbor evidence for
temporal completion and learns the remaining benefit of new
collaboration to guide evidence acquisition and fusion.

\begin{figure*}[t]
  \centering
  \includegraphics[width=\textwidth,clip]{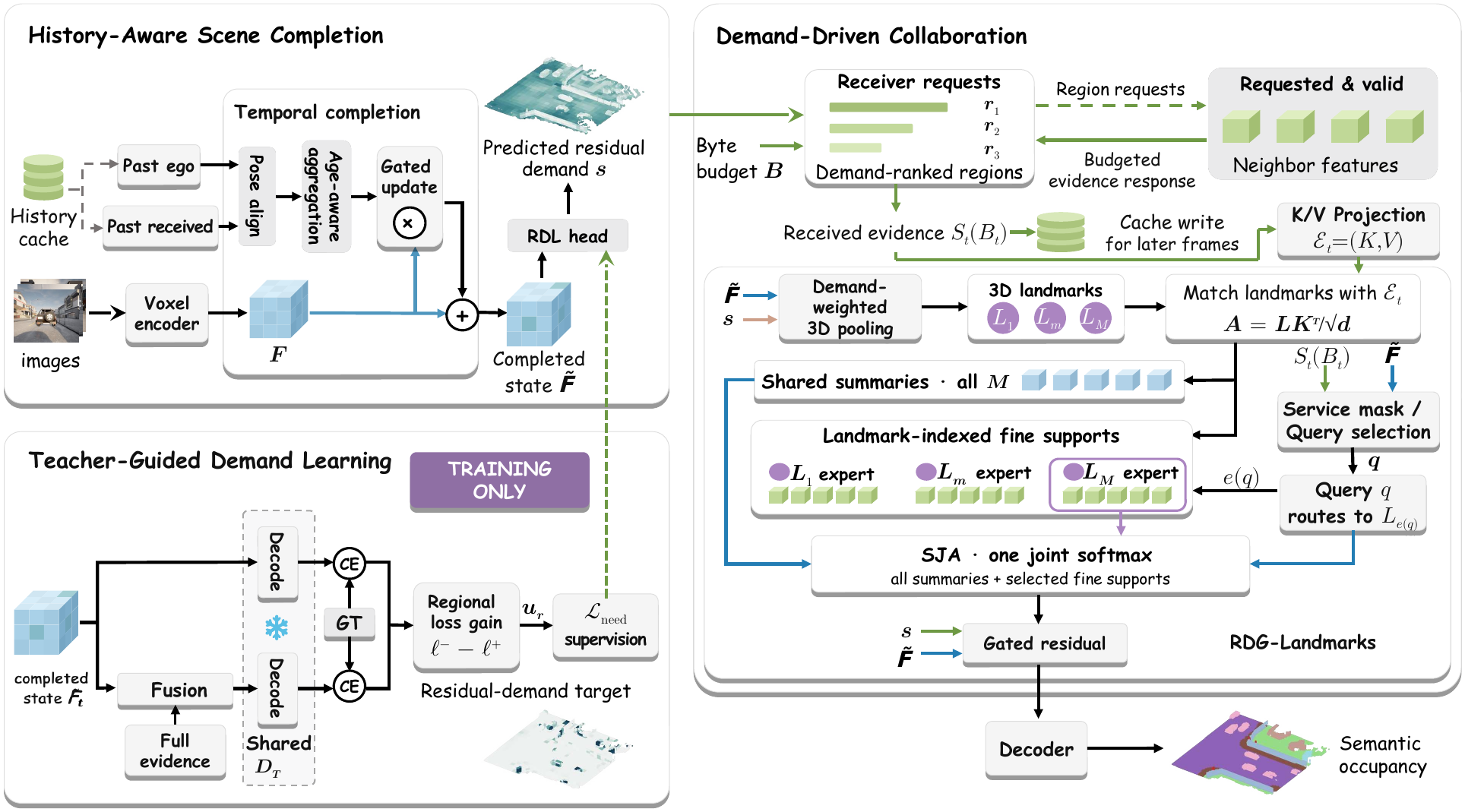}
  \caption{Overview of \method. Temporal completion combines
  current ego features with aligned history and received observations.
  Remaining demand guides budgeted requests and weighted 3D landmarks.
  Shared summaries and fine supports provide reusable evidence for
  query-specific fusion before occupancy decoding. The cache branch stores
  decoded received features and acquisition metadata for later frames;
  the highlighted expert illustrates one query-dependent route.
  The teacher branch
  supplies regional loss-gain targets during training.}
  \label{fig:framework}
\end{figure*}
\section{Method}
\label{sec:method}

\subsection{\texorpdfstring{Framework Overview}{Framework Overview}}
\label{sec:setup}

At time $t$, ego $e$ and neighbors $j\in\mathcal N_t$ observe
multi-view images. We reuse VQSOP's encoder $E$ and occupancy
decoder $D$~\cite{li2026vqsop}, replacing its cross-agent fusion.
Before quantization, the encoder produces voxel features
$F_t^j\in\mathbb R^{C\times Z\times H\times W}$,
with $C$ channels and spatial dimensions $Z,H,W$.
Given a communication budget $B_t$, \method first completes
ego features using historical observations, then predicts
remaining collaboration demand through teacher-supervised
RDL. The predicted demand guides budgeted evidence acquisition
and RDG-Landmarks construction, while SJA integrates received
evidence for occupancy decoding (Fig.~\ref{fig:framework}).

\subsection{History-Aware Scene Completion}
\label{sec:temporal}

\paragraph{Pose-aligned historical aggregation.}
We maintain detached historical ego features with acquisition
poses and previously received, decoded neighbor tokens with source
identities, timestamps, and world coordinates.
Let $T^w_{j,\tau}$ map source-local coordinates to the
world frame. Historical ego features from the last $T_h$ frames
are aligned and aggregated:
\begin{equation}
\begin{aligned}
T_{e_t\leftarrow j_\tau}
&=(T^w_{e,t})^{-1}T^w_{j,\tau},\\
H_t(v)
&=\frac{\sum_{\tau<t}w_{\tau,t}(v)
\mathcal W(F_\tau^e,T_{e_t\leftarrow e_\tau})(v)}
{\sum_{\tau<t}w_{\tau,t}(v)+\epsilon},
\end{aligned}
\label{eq:history}
\end{equation}
where $\mathcal W$ resamples metric voxel centers and
$w_{\tau,t}(v)=m_{\tau,t}(v)\exp(-(t-\tau)/\tau_h)$ uses a geometric
validity mask $m_{\tau,t}$ and decay time scale $\tau_h$; $v$ indexes voxels
and $\epsilon>0$ stabilizes normalization. Cached neighbor tokens are transformed into the current
ego frame and scattered into $C_t$, retaining support and age information.

\paragraph{Gated temporal completion.}
Pointwise gate $g_h$ and feature branch $\phi_h$ read context $z_t$
comprising current features, their differences from $H_t$ and $C_t$, support
masks, and observation ages. The completed features are
\begin{equation}
\widetilde F_t=F_t^e+
 m_t^h\odot\sigma(g_h(z_t))\odot W_h\phi_h(z_t),
\label{eq:temporal}
\end{equation}
where $m_t^h$ marks available history, $\sigma$ is the sigmoid function, and
$\odot$ denotes elementwise multiplication. The output projection $W_h$ is
zero-initialized to preserve the ego path initially. Without available
history, the residual update is zero.

\subsection{Teacher-Guided Demand Learning}
\label{sec:rdl}
\label{sec:demand}

\paragraph{Residual collaboration demand.}
We define residual collaboration demand as the potential
occupancy improvement achievable from new neighbor evidence
after temporal completion. Given the completed features
$\widetilde F_t$ and temporal context $z_t$, the demand
predictor outputs
\begin{equation}
s_t(v)=\operatorname{softplus}
\bigl(h_{\rm need}(\widetilde F_t(v),z_t(v))\bigr).
\label{eq:demand_prediction}
\end{equation}
\paragraph{Teacher-supervised regional gain.}
The teacher compares predictions from the same completed ego state
$\widetilde F_t^T$ and decoder $D_T$, differing in whether new collaborative
evidence is available:
\begin{equation}
\begin{aligned}
p_T^-&=D_T(\widetilde F_t^T),\\
p_T^+&=D_T\bigl(\mathcal F_T(\widetilde F_t^T,S_t^{\rm all})\bigr).
\end{aligned}
\label{eq:teacher_predictions}
\end{equation}
Here $S_t^{\rm all}$ contains all available source evidence and
$\mathcal F_T$ is teacher fusion, using the same payload precision as the
budgeted path. For valid labeled output voxels $\mathcal V_r$ in
communication region $r$, the target is
\begin{equation}
u_r=\left[
\frac{1}{|\mathcal V_r|}
\sum_{v\in\mathcal V_r}
\bigl(\ell(p_T^-(v),y_v)-\ell(p_T^+(v),y_v)\bigr)
\right]_+,
\label{eq:gain}
\end{equation}
where $\ell$ is cross-entropy with optional class weights,
and $[x]_+=\max(x,0)$.
Regions without valid labels are excluded.
Averaging before truncation measures the net regional
improvement, while the shared completed state isolates
the additional benefit of new collaborative evidence.

\paragraph{Demand supervision and inference.}
We broadcast $u_r$ to its corresponding latent region and train
the demand predictor with $\mathcal L_{\rm need}$, combining Smooth-$L_1$
regression for gain magnitude and within-frame pairwise ranking for
communication priorities.
Targets are detached, and empty supervision sets contribute zero.
Supplementary Sec.~A provides the full loss and target construction.
The teacher supplies supervision only: inference obtains $s_t$ directly
from the demand predictor, without ground truth or full-information access.

\subsection{\texorpdfstring{Demand-Driven Collaboration}{Demand-Driven Collaboration}}
\label{sec:collaboration}
\label{sec:rdg_landmarks}

\paragraph{Demand-prioritized evidence acquisition.}
We rank disjoint communication regions $\{\mathcal R_r\}$
by their mean predicted demand. Requests specify region indices and
priorities, while senders return geometrically valid features from the
requested regions. A deterministic packer admits request--response
exchanges in priority order under the byte budget:
\begin{equation}
\sum_{p\in\mathcal P_t}
\bigl(\operatorname{bytes}(p^{\rm req})+
      \operatorname{bytes}(p^{\rm resp})\bigr)\leq B_t.
\label{eq:communication_budget}
\end{equation}
Here $\mathcal P_t$ denotes admitted exchanges, and the byte count
includes FP16 features, indices, identities, timestamps, and geometry. Request
priorities use only receiver-side information, and only decoded responses
enter fusion.

Let $S_t(B_t)$ contain $N_s$ decoded received tokens. Regions containing
valid returned tokens are serviced; $\mathcal Q_t$ contains $N_q$ queries
extracted from $\widetilde F_t$ at voxel locations in these regions.
Budget changes affect admitted evidence and serviced regions, while $M$,
$k$, history length, and model weights remain fixed. Fine-support selection
operates on received tokens without additional transmission.

\paragraph{Demand-weighted 3D landmarks.}
Fixed spatial bins provide a structured representation, but uniform pooling
does not distinguish locations with different remaining needs.
We project $\widetilde F_t$ to $U$ and partition the ego volume into $M$
nonoverlapping 3D bins $\mathcal B_m$. Writing $s_v=s_t(v)$, we normalize
demand by a stop-gradient frame mean,
$\bar s_v=s_v/(\operatorname{sg}(\operatorname{mean}_v s_v)+\epsilon)$,
and construct
\begin{equation}
L_m=
\frac{\sum_{v\in\mathcal B_m}\nu_v(\epsilon_l+\bar s_v)U_v}
     {\sum_{v\in\mathcal B_m}\nu_v(\epsilon_l+\bar s_v)}.
\label{eq:landmarks}
\end{equation}
Here $\nu_v$ combines current, historical, and cached support, and
$\epsilon_l>0$ preserves contextual contributions from low-demand locations.
Empty bins are masked; blind regions retain finite ego features and remain
eligible for requests. Constant demand weights recover uniform pooling.
The bins remain fixed, while their representatives adapt to demand,
linking evidence acquisition to subsequent fusion.

\paragraph{Shared summaries and fine supports.}
Summaries provide compact context but can lose local detail. We retain
both pooled summaries and selected original features by adapting
shared-support attention~\cite{wen2026mita} to ego-query and received-evidence
sets of potentially different sizes. For one attention head of dimension
$d$, let $Q$ denote projected ego queries and $\mathcal E_t=(K,V)$ denote
the keys and values projected from the decoded received tokens $S_t(B_t)$.
Let $L$ denote the projected demand-weighted landmarks.
Positions condition queries and keys; source age also conditions keys.
We compute
\begin{equation}
\begin{aligned}
A&=LK^\top/\sqrt d,\\
\overline V&=\operatorname{softmax}(A)V,\qquad
\mathcal I_m=\operatorname{TopK}(A_m,k).
\end{aligned}
\label{eq:experts}
\end{equation}
The $M$ summaries aggregate all valid received evidence, while
$\mathcal I_m$ retains at most $k$ original neighbor keys and values for each
landmark and head. Selecting these fine supports once per landmark makes
them reusable across queries without replacing their original content
with pooled features.

\paragraph{Shared-support joint fusion.}
\label{sec:sja}
Each query $q$ routes to $L_{e(q)}$, where
$e(q)=\arg\max_m qL_m^\top$, and accesses all summaries together with that
expert's fine supports:
\begin{equation}
\begin{aligned}
K_q^*&=[L;K_{\mathcal I_{e(q)}}],\\
V_q^*&=[\overline V;V_{\mathcal I_{e(q)}}],
\end{aligned}
\label{eq:query_supports}
\end{equation}
where $[\,;\,]$ denotes concatenation. Landmarks serve as summary keys paired
with $\overline V$. Shared summaries and original fine tokens participate
in a single attention normalization:
\begin{equation}
o_q=\operatorname{softmax}
\left(qK_q^{*\top}/\sqrt d\right)V_q^*.
\label{eq:joint}
\end{equation}
This joint readout lets each query distribute attention between summarized
context and unpooled detail. Queries assigned to the same expert share
support indices, not attention weights or outputs.

The concatenated head outputs $o_v$ are written back through a
demand-conditioned residual gate:
\begin{equation}
\begin{aligned}
F_t^{\rm out}(v)={}&\widetilde F_t(v)\\
&+\chi_v\,\sigma(g_c([U_v,s_v]))\,W_c o_v.
\end{aligned}
\label{eq:residual}
\end{equation}
The mask $\chi_v$ combines query membership in $\mathcal Q_t$ with evidence
availability; without usable received evidence, the collaborative update
is zero. The gate regulates how
strongly received evidence modifies the completed state, and the
zero-initialized projection $W_c$ preserves that state initially.
The decoder $D$ maps $F_t^{\rm out}$ to semantic occupancy predictions.

\subsection{Training Strategy and Objectives}
\label{sec:training}

\paragraph{Teacher preparation and budgeted training.}
With the pretrained encoder frozen, we warm up temporal completion and
full-information fusion by supervising both completed-ego and fused
predictions. Historical source memory uses actual budgeted past messages,
so full-information access does not enter through the history cache.
We then freeze the teacher, its decoder, and the warmed-up temporal module.
Teacher and student use consistent temporal conditioning, keeping the
regional gain target anchored to the same completed ego state.
Budgeted collaboration is trained with
\begin{equation}
\mathcal L=\mathcal L_{\rm occ}
+\lambda_h\mathcal L_{\rm pre}
+\lambda_n\mathcal L_{\rm need}.
\label{eq:total_loss}
\end{equation}
Here $\mathcal L_{\rm occ}$ supervises fused occupancy predictions,
$\mathcal L_{\rm pre}$ supervises completed-ego predictions, and
$\mathcal L_{\rm need}$ supervises residual demand. The occupancy losses
combine cross-entropy with semantic and geometric scaling losses;
$\lambda_h$ and $\lambda_n$ balance the objectives.
Supplementary Sec.~A expands $\mathcal L_{\rm need}$.
Ground truth and full-information teacher access are restricted to training
and offline diagnostics.

\hyphenation{CoHFF VQSOP mIoU IoU Occ KiB}
\section{Experiments}
\label{sec:experiments}

\paragraph{Dataset and evaluation.}
\label{sec:setup_experiments}
We evaluate the same 1,980 SemanticOPV2V validation frames~\cite{song2024cohff,xu2022opv2v}
using semantic mIoU and occupied-space IoU (Occ.~IoU), excluding unknown
labels. Occ.~IoU merges occupied semantic categories. Gap2Co reuses VQSOP's
Stage-1 image-to-voxel encoder and decoding interface~\cite{li2026vqsop}.
Our main model uses ego-history length $T_h=4$, $M=50$ landmarks,
and fine-support size $k=8$. Communication is measured in KiB/frame
per receiver, counting serialized requests and responses across all
neighbors ($1\,\mathrm{KiB}=1024$ bytes);
accuracy differences are in percentage points.
Tables~\ref{tab:main}--\ref{tab:ablations} report means over five training
seeds. Corresponding curves reuse and average the same checkpoints.

\subsection{Main Results}
\label{sec:main_results}

\begin{table*}[t]
\centering
\small
\setlength{\tabcolsep}{3.0pt}
\renewcommand{\arraystretch}{1.08}
\caption{Semantic occupancy comparison (\%). All entries are
averaged over five independent training runs per method. $\dagger$: Our single-agent variant uses the local
ego-temporal model with zero neighbor communication.
Both settings use 1,980 existing-label validation frames. Best and second-best in each
setting are bold and underlined.}
\label{tab:main}
\begin{tabular*}{\textwidth}{@{\extracolsep{\fill}}*{9}{c}@{}}
\toprule
\multirow{2}{*}{Metric / Class} & \multicolumn{4}{c}{Single-Agent} & \multicolumn{4}{c}{Collaborative Perception} \\
\cmidrule(lr){2-5}\cmidrule(lr){6-9}
 & CoHFF~\cite{song2024cohff} & GSFormer~\cite{huang2024gaussianformer} & VQSOP~\cite{li2026vqsop} & \textbf{Ours}$^{\dagger}$ & CoHFF~\cite{song2024cohff} & GSFusion~\cite{chen2026vogs} & VQSOP~\cite{li2026vqsop} & \textbf{Ours} \\
\midrule
Occ.~IoU & 39.02 & 67.24 & \underline{70.92} & \textbf{76.90} & 50.87 & 72.64 & \underline{73.34} & \textbf{80.95} \\
mIoU & 25.27 & 28.74 & \underline{34.08} & \textbf{40.46} & 34.59 & 37.27 & \underline{41.18} & \textbf{45.74} \\
\midrule
Building & \textbf{21.41} & 3.49 & 10.59 & \underline{11.32} & \textbf{26.26} & 9.41 & 10.81 & \underline{11.15} \\
Fence & 21.11 & 13.43 & \textbf{30.87} & \underline{25.65} & 28.26 & 28.77 & \underline{32.87} & \textbf{34.13} \\
Terrain & 44.45 & \underline{68.49} & 48.78 & \textbf{69.96} & 48.81 & \textbf{74.26} & 62.70 & \underline{68.17} \\
Pole & \textbf{31.94} & 5.74 & \underline{12.73} & 12.02 & \textbf{43.03} & 12.35 & \underline{19.91} & 18.00 \\
Road & 56.27 & \underline{78.73} & 76.01 & \textbf{84.46} & 62.45 & 82.67 & \underline{86.68} & \textbf{88.04} \\
Sidewalk & 18.06 & \underline{69.77} & 58.52 & \textbf{72.88} & 39.94 & \textbf{77.88} & 66.90 & \underline{77.69} \\
Vegetation & 14.81 & 12.25 & \underline{23.64} & \textbf{27.85} & 21.14 & 20.50 & \underline{26.61} & \textbf{30.53} \\
Vehicles & \textbf{59.21} & 48.40 & 52.99 & \underline{53.11} & \underline{63.75} & 60.19 & 62.93 & \textbf{68.30} \\
Wall & 33.86 & 30.06 & \underline{35.45} & \textbf{38.47} & \textbf{58.63} & 36.17 & 36.27 & \underline{40.52} \\
Guardrail & 2.12 & 14.52 & \textbf{29.10} & \underline{26.02} & 2.07 & 32.63 & \textbf{53.81} & \underline{51.23} \\
Traffic sign & 0.00 & 0.00 & \underline{13.87} & \textbf{19.09} & \textbf{16.78} & 8.15 & \underline{13.73} & 12.83 \\
Bridge & 0.00 & 0.00 & \underline{16.41} & \textbf{44.72} & 3.96 & 4.26 & \underline{20.94} & \textbf{48.33} \\
\bottomrule
\end{tabular*}
\end{table*}

\paragraph{Single-agent perception.}
In Table~\ref{tab:main}, Gap2Co (Ego-only) achieves 40.46\% mIoU and
76.90\% Occ.~IoU without communication, exceeding VQSOP by 6.38 and
5.98 points. Road/sidewalk IoUs reach 84.46\%/72.88\%, versus
76.01\%/58.52\%. These gains demonstrate that ego temporal completion
provides a strong basis for collaboration.

\paragraph{Full-model performance.}
Full Gap2Co reaches \textbf{45.74\% mIoU and 80.95\% Occ.~IoU}, exceeding
collaborative VQSOP by 4.56 and 7.61 points and Ego-only by 5.28 and
4.05 points. Both metrics demonstrate that demand-driven
collaboration improves semantics and occupied-space reconstruction beyond
the completed ego representation.

\subsection{Ablation Studies}
\label{sec:ablations}

Table~\ref{tab:selection_fusion} compares selection and aggregation,
while Table~\ref{tab:ablations} activates inference components cumulatively.
In Table~\ref{tab:selection_fusion},
cooperative variants use $M=50$, $k=8$, $T_h=4$, and a reference budget of
54.24~KiB/frame; ego-only communication is zero.
In Table~\ref{tab:ablations}, all variants share fixed weights within each
checkpoint. A/A0 communicate zero bytes; B--E use a 64~KiB/frame cap
(Full: 54.24~KiB/frame on average).

\begin{table}[t]
\centering
\small
\setlength{\tabcolsep}{3.5pt}
\renewcommand{\arraystretch}{1.04}
\caption{Information selection and evidence aggregation (\%).
(A) Selection with Uniform-Landmark MiTA, fine supports enabled,
and neighbor reuse disabled. (B) Aggregation under Remaining Demand selection, with identical
received evidence and fine supports, but without neighbor reuse. (C) Request-ranking and reuse controls with
Demand-weighted MiTA. Best scores within each block are bold.}
\label{tab:selection_fusion}
\label{tab:demand}
\label{tab:fusion}
\begin{tabular}{@{}>{\raggedright\arraybackslash}p{0.66\columnwidth}rr@{}}
\toprule
Method & mIoU$\uparrow$ & Occ.~IoU$\uparrow$ \\
\midrule
Ego-only + ego history & 40.46 & 76.90 \\
\midrule
\multicolumn{3}{@{}l}{\textit{A. Selection criterion}} \\
Random & 41.89 & 77.46 \\
Confidence & 42.34 & 77.11 \\
Temporal difference & 42.12 & 77.53 \\
Feature magnitude & 42.21 & 77.38 \\
Entropy & 42.72 & 77.94 \\
\textbf{Remaining Demand} & \textbf{43.38} & \textbf{79.12} \\
\midrule
\multicolumn{3}{@{}l}{\textit{B. Evidence aggregation (Demand selection)}} \\
Flattened cross-attention & 41.53 & 77.86 \\
Uniform-Landmark MiTA & 43.38 & 79.12 \\
Demand-weighted MiTA (no reuse) & \textbf{44.37} & \textbf{79.63} \\
\midrule
\multicolumn{3}{@{}l}{\textit{C. Demand and history controls}} \\
Entropy + Demand-weighted (no reuse) & 43.35 & 78.53 \\
Entropy + Demand-weighted (reuse) & 44.71 & 80.11 \\
\textbf{Gap2Co (Full, Demand + reuse)} & \textbf{45.74} & \textbf{80.95} \\
\bottomrule
\end{tabular}
\end{table}

\paragraph{Communication-region selection.}
\label{sec:demand_experiments}
Table~\ref{tab:selection_fusion}(A) changes only region ranking, fixing the
completed ego state, candidates, protocol, and architecture. Remaining
Demand achieves 43.38\% mIoU and 79.12\% Occ.~IoU, exceeding
Entropy by 0.66 and 1.18 points. This supports residual-gain
estimation after temporal completion as a more effective selection signal
than prediction uncertainty.

\paragraph{Cooperative evidence organization.}
\label{sec:fusion_experiments}
With identical Remaining Demand selection, received evidence, and history,
Uniform-Landmark MiTA improves over flattened cross-attention by
1.85 mIoU points (Table~\ref{tab:selection_fusion}(B)). Changing only
landmark weights, with fine supports and SJA fixed, raises mIoU from
43.38\% to 44.37\% (+0.99). The two comparisons isolate
evidence organization and demand-weighted pooling, respectively.

\paragraph{Unified demand guidance and historical reuse.}
Within Demand-weighted MiTA, Remaining Demand exceeds Entropy by
1.02 mIoU points without reuse and by 1.03 mIoU/0.84
Occ.~IoU points with reuse (Table~\ref{tab:selection_fusion}(C)). Reuse adds
1.37 mIoU points to demand-guided fusion. These controls show that
remaining-demand selection and historical reuse provide complementary gains.

\begin{table}[t]
\centering
\small
\setlength{\tabcolsep}{2pt}
\renewcommand{\arraystretch}{1.04}
\caption{Cumulative ablation averaged over five training checkpoints.
$\Delta$: mean paired mIoU gain over the preceding row.
A: $T_h=0$; A0--E: $T_h=4$. A/A0 are ego-only.
B--E use Remaining Demand for region
selection. B--C use landmark summaries only, while D--E
add fine supports; E further enables historical neighbor reuse.}
\label{tab:ablations}
\begin{tabular}{@{}l>{\raggedright\arraybackslash}p{0.44\columnwidth}rrr@{}}
\toprule
ID & Added component & mIoU$\uparrow$ & Occ.~IoU$\uparrow$ & $\Delta$ mIoU \\
\midrule
A & Single-frame ego-only & 38.06 & 74.38 & -- \\
A0 & + Ego history & 40.46 & 76.90 & +2.40 \\
B & + Current-neighbor summaries (Uniform LM) & 41.03 & 77.39 & +0.57 \\
C & + Demand-weighted pooling & 42.89 & 78.56 & +1.86 \\
D & + Landmark-specific fine supports & 44.37 & 79.63 & +1.48 \\
\textbf{E} & \textbf{+ Historical neighbor reuse (Full)} & \textbf{45.74} & \textbf{80.95} & \textbf{+1.37} \\
\bottomrule
\end{tabular}
\end{table}

\paragraph{Cumulative Component Ablation.}

With fixed weights (Table~\ref{tab:ablations}), ego history raises
mIoU from 38.06\% to 40.46\% (+2.40). Current-neighbor summaries further
improve it to 41.03\% (+0.57). With identical evidence,
demand-weighted pooling achieves 42.89\% (+1.86), followed by
44.37\% (+1.48) with fine supports at no extra communication cost.
Historical neighbor reuse further raises mIoU to 45.74\% (+1.37).
Together, these components deliver gains of 5.28 mIoU and
4.05 Occ.~IoU points over Ego-only, demonstrating the effectiveness
of demand-guided evidence acquisition and fusion.

\begin{figure}[!t]
\centering
\begin{subfigure}[t]{0.49\columnwidth}
\centering
\includegraphics[width=\linewidth,clip]{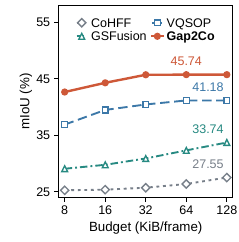}
\caption{Semantic mIoU}\label{fig:budget_miou}
\end{subfigure}\hfill
\begin{subfigure}[t]{0.49\columnwidth}
\centering
\includegraphics[width=\linewidth,clip]{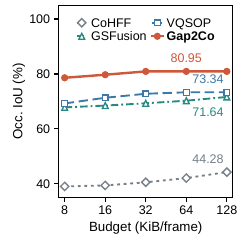}
\caption{Occupancy IoU}\label{fig:budget_occ}
\end{subfigure}
\caption{Accuracy under communication-budget constraints.
(a) Semantic mIoU and (b) occupancy IoU as the budget increases from 8 to
128~KiB per ego frame. The cap covers all senders to the same receiving
ego and counts all request and response bytes exchanged with its neighbors,
with $1\,\mathrm{KiB}=1024$ bytes.}
\label{fig:budget}
\end{figure}

\begin{figure}[!t]
\centering
\begin{subfigure}[t]{0.49\columnwidth}
\centering
\includegraphics[width=\linewidth,clip]{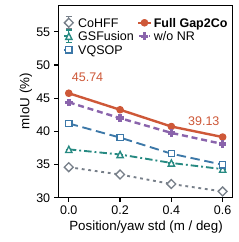}
\caption{Pose errors}\label{fig:robustness_pose}
\end{subfigure}\hfill
\begin{subfigure}[t]{0.49\columnwidth}
\centering
\includegraphics[width=\linewidth,clip]{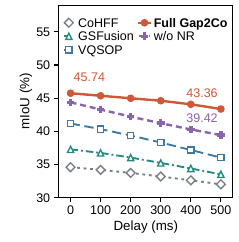}
\caption{Communication delay}\label{fig:robustness_delay}
\end{subfigure}
\caption{Robustness to pose errors and communication delay.
Both panels use an unlimited communication budget.
(a) Coupled Gaussian $x/y$ translation and yaw errors, with paired standard deviations
from $0$~m/$0^\circ$ to $0.6$~m/$0.6^\circ$. This panel reports mIoU on
1,980 validation frames; nonzero pose settings are averaged over five perturbation
seeds shared across methods.
(b) Fixed delays from 0 to 500~ms, with identical time offsets across methods.
The w/o NR curve denotes Full Gap2Co w/o Neighbor Reuse.}
\label{fig:robustness}
\end{figure}

\begin{figure*}[!tp]
\centering
\includegraphics[width=\textwidth,clip]{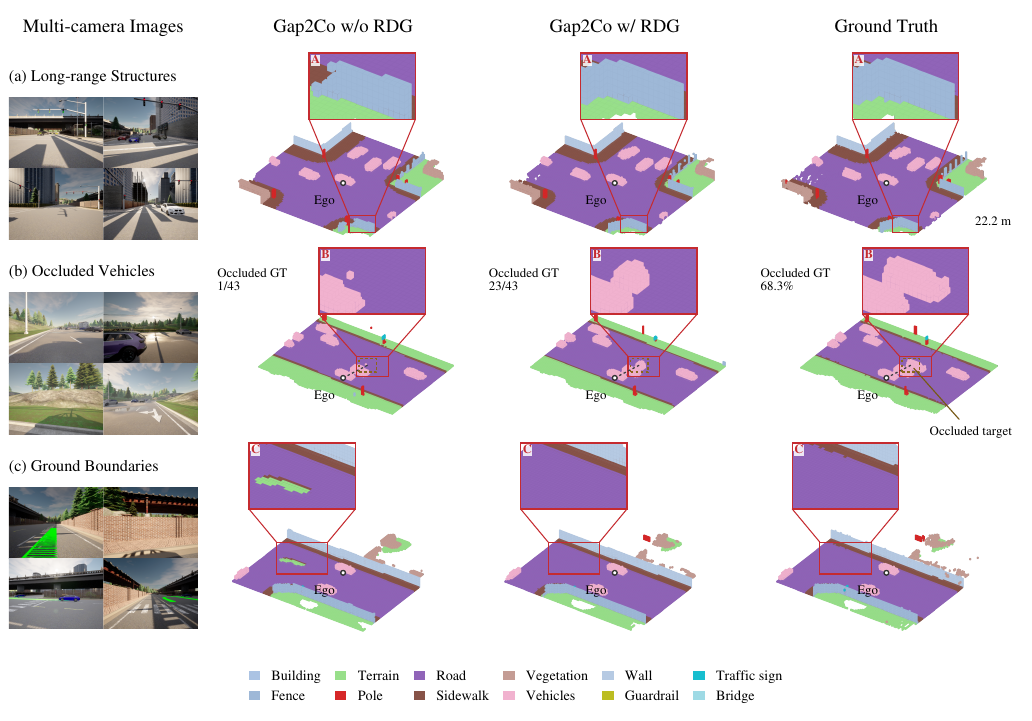}
\caption{Qualitative occupancy comparisons. Rows show
(a) long-range structures, (b) occluded vehicles, and (c) ground boundaries.
``w/o RDG'' disables the entire RDG-Landmarks module;
``w/ RDG'' uses the full module.
Multi-camera images, predictions, and ground truth are shown for each
scene. Red boxes and enlarged views highlight the local geometry;
voxel colors indicate semantic classes.}
\label{fig:qualitative}
\end{figure*}

\subsection{Robustness Analysis}
\label{sec:robustness}

We evaluate robustness to communication budgets, pose errors,
delays, and occlusion using fixed weights and matched
evaluation samples within each setting.
Compared methods share external perturbation realizations;
pairing details are provided in the supplementary material.

\paragraph{Communication-budget analysis.}
Figure~\ref{fig:budget} uses matched
neighbors and 8--128~KiB/frame request--response caps without an additional
neighbor-count limit. Gap2Co achieves 42.67\% mIoU/78.62\% Occ.~IoU at
8~KiB, 45.72\%/80.94\% at 32~KiB, and 45.74\%/80.95\% at 64--128~KiB.
Thus most collaborative benefit is obtained at low communication budgets.
GSFusion and CoHFF improve with larger budgets but remain below their
full-communication references at 128~KiB (Table~\ref{tab:main});
strict caps can discard Gaussian or feature evidence.
At 64~KiB, Gap2Co exceeds VQSOP, GSFusion, and CoHFF by 4.56, 13.39, and
19.34 mIoU points. These gains support demand-driven acquisition and fusion
of evidence that remains useful after temporal completion.

\paragraph{Pose-error robustness.}
Under unlimited communication (Fig.~\ref{fig:robustness}\subref{fig:robustness_pose}),
\emph{Full Gap2Co w/o Neighbor Reuse} (w/o NR) retains demand-weighted
aggregation and fine supports, scoring 44.37\% mIoU at zero noise, as in
Table~\ref{tab:ablations}, row D. Accuracy falls with noise, but w/o NR remains
above VQSOP and below Full Gap2Co. At $0.6$~m/$0.6^\circ$, Full Gap2Co
and w/o NR achieve \textbf{39.13\%} and \textbf{38.08\% mIoU}, versus
34.96\% for VQSOP, demonstrating the benefit of historical reuse under
spatial misalignment.

\paragraph{Communication-delay robustness.}
For 0--500~ms delays under the same unlimited budget
(Fig.~\ref{fig:robustness}\subref{fig:robustness_delay}), Full Gap2Co falls
from 45.74\% to \textbf{43.36\% mIoU} (\textbf{2.38 points}), whereas
w/o NR falls from 44.37\% to \textbf{39.42\%} (\textbf{4.95 points}).
The \textbf{3.94-point} advantage at 500~ms supports historical evidence
reuse for mitigating information staleness.

\paragraph{Occlusion and complementary neighbor evidence.}
Ground-truth occupancy, depth maps, and camera calibration determine
visibility for offline grouping. Let $\mathcal V_e$ and $\mathcal U_e$ denote
directly visible and confirmed occluded occupied-surface voxels in the ego
field of view. Ego occlusion and complementary neighbor evidence are
\begin{equation}
\begin{aligned}
r_{\mathrm{occ}} &= \frac{|\mathcal U_e|}{|\mathcal V_e|+|\mathcal U_e|},\\
\mathrm{NE} &= \frac{|\mathcal U_e\cap(\bigcup_j\mathcal V_j)|}{|\mathcal U_e|},
\end{aligned}
\label{eq:occlusion_ne}
\end{equation}
where $\mathcal V_j$ is the occupied-surface set visible to neighbor $j$.
Among 1,979 frames with at least 32 confirmed ego-occluded voxels, global
occlusion tertiles are 21.78\%/32.04\% and median NE is 43.05\%.
We fix these groups, then select 990 frames with $\mathrm{NE}\geq43.05\%$.
Both methods use identical frames per row and pool full-grid confusion counts.
\begin{table}[t]
\centering
\small
\setlength{\tabcolsep}{2.75pt}
\renewcommand{\arraystretch}{1.10}
\caption{mIoU by ego occlusion with complementary neighbor evidence.
Only frames with $\mathrm{NE}\geq43.05\%$ are shown (990 total).
NE is the fraction of ego-occluded occupied voxels visible to at least
one neighbor. Scores pool full-grid confusion counts within each row.
Occlusion bins are global tertiles of all 1,979 eligible
frames; thresholds are rounded for display. $\Delta$ is Gap2Co minus VQSOP
in percentage points (pp).}
\label{tab:occlusion}
\begin{tabular}{@{}lcrrrr@{}}
\toprule
\multicolumn{2}{c}{Ego occlusion} & \multirow{2}{*}{Frames} &
\multicolumn{2}{c}{mIoU (\%)} & \multirow{2}{*}{$\Delta$ (pp)} \\
\cmidrule(lr){1-2}\cmidrule(lr){4-5}
Level & $r_{\mathrm{occ}}$ (\%) & & VQSOP & Gap2Co & \\
\midrule
Low & $<21.78$ & 359 & 42.36 & \textbf{48.21} & +5.85 \\
Medium & $[21.78,32.04)$ & 243 & 42.28 & \textbf{44.02} & +1.74 \\ 
High & $\geq32.04$ & 388 & 35.03 & \textbf{39.02} & +3.99 \\
\bottomrule
\end{tabular}
\end{table}
Within these high-NE subsets (Table~\ref{tab:occlusion}), Gap2Co improves mIoU over VQSOP by
5.85, 1.74, and 3.99 points in the low-, medium-, and high-occlusion groups,
respectively. On the 388 high-occlusion frames with complementary neighbor
coverage, it reaches \textbf{39.02\% mIoU}, compared with 35.03\% for VQSOP.
These results highlight the effectiveness of demand-driven
collaboration in exploiting complementary neighbor evidence
under ego occlusion.

\subsection{Qualitative Visualization}
\label{sec:qualitative}
Figure~\ref{fig:qualitative}
compares Gap2Co with RDG-Landmarks enabled and disabled.
With the module enabled, (a) the structure at 22.2~m is more complete;
(b) recovered occupied voxels of the partially occluded vehicle increase
from 1/43 to 23/43; and (c) the road--sidewalk boundary follows the
ground-truth geometry more closely, demonstrating improved completion
of structures and occluded objects and refinement of semantic boundaries.

\section{Conclusion}
\label{sec:conclusion}

We presented \method for collaborative 3D semantic occupancy prediction.
It integrates temporal completion, teacher-guided residual demand learning,
and demand-driven acquisition and fusion with RDG-Landmarks and SJA.
Experiments demonstrate improved occupancy accuracy and communication
efficiency, validating history-aware scene completion,
teacher-guided residual demand learning, and demand-driven collaboration.
Future work will explore adaptive cooperative memory
management to improve efficiency and robustness in dynamic scenes.
\clearpage
\section*{LLM Usage Statement}
ChatGPT and Codex (OpenAI) were used to generate and revise materials
for manuscript writing and editing, related work and citations,
code and implementation, experimental design and execution,
analysis of results, figures and visualizations, data preparation,
and the development of research ideas, hypotheses, claims, and conclusions.
These uses involved one-shot or interactive generation, with the authors
reviewing, editing, or approving the generated outputs.
The authors verified the final content and take full responsibility
for all submitted content, including the methods, data, code, results,
and conclusions.


{\small
\begin{thebibliography}{22}
\providecommand{\natexlab}[1]{#1}
\providecommand{\url}[1]{\texttt{#1}}
\expandafter\ifx\csname urlstyle\endcsname\relax
  \providecommand{\doi}[1]{doi: #1}\else
  \providecommand{\doi}{doi: \begingroup \urlstyle{rm}\Url}\fi

\bibitem[Chen et~al.(2026)Chen, Huang, and Bagchi]{chen2026vogs}
Cheng Chen, Hao Huang, and Saurabh Bagchi.
\newblock Vision-only gaussian splatting for collaborative semantic occupancy
  prediction.
\newblock \emph{Proceedings of the AAAI Conference on Artificial Intelligence},
  40\penalty0 (4):\penalty0 2796--2804, 2026.

\bibitem[Hong et~al.(2024)Hong, Liu, Li, Li, and He]{hong2024mrcnet}
Shixin Hong, Yu Liu, Zhi Li, Shaohui Li, and You He.
\newblock Multi-agent collaborative perception via motion-aware robust
  communication network.
\newblock In \emph{Proceedings of the IEEE/CVF Conference on Computer Vision
  and Pattern Recognition}, pages 15301--15310, 2024.

\bibitem[Hu et~al.(2022)Hu, Fang, Lei, Zhong, and Chen]{hu2022where2comm}
Yue Hu, Shaoheng Fang, Zixing Lei, Yiqi Zhong, and Siheng Chen.
\newblock {Where2comm}: Communication-efficient collaborative perception via
  spatial confidence maps.
\newblock In \emph{Advances in Neural Information Processing Systems}, 2022.

\bibitem[Hu et~al.(2024)Hu, Peng, Liu, Ge, Liu, and Chen]{hu2024codefilling}
Yue Hu, Juntong Peng, Sifei Liu, Junhao Ge, Si Liu, and Siheng Chen.
\newblock Communication-efficient collaborative perception via information
  filling with codebook.
\newblock In \emph{Proceedings of the IEEE/CVF Conference on Computer Vision
  and Pattern Recognition}, pages 15481--15490, 2024.

\bibitem[Hu et~al.(2026)Hu, Pang, Qin, Eldar, Chen, Zhang, and
  Zhang]{hu2026pragcomm}
Yue Hu, Xianghe Pang, Xiaoqi Qin, Yonina~C. Eldar, Siheng Chen, Ping Zhang, and
  Wenjun Zhang.
\newblock Pragmatic communication in multi-agent collaborative perception.
\newblock \emph{IEEE Transactions on Pattern Analysis and Machine
  Intelligence}, 48\penalty0 (8):\penalty0 9279--9296, 2026.

\bibitem[Huang et~al.(2023)Huang, Zheng, Zhang, Zhou, and Lu]{huang2023tpv}
Yuanhui Huang, Wenzhao Zheng, Yunpeng Zhang, Jie Zhou, and Jiwen Lu.
\newblock Tri-perspective view for vision-based {3D} semantic occupancy
  prediction.
\newblock In \emph{Proceedings of the IEEE/CVF Conference on Computer Vision
  and Pattern Recognition}, pages 9223--9232, 2023.

\bibitem[Huang et~al.(2024)Huang, Zheng, Zhang, Zhou, and
  Lu]{huang2024gaussianformer}
Yuanhui Huang, Wenzhao Zheng, Yunpeng Zhang, Jie Zhou, and Jiwen Lu.
\newblock {GaussianFormer}: Scene as gaussians for vision-based {3D} semantic
  occupancy prediction.
\newblock In \emph{European Conference on Computer Vision}, 2024.

\bibitem[Leng et~al.(2025)Leng, Yang, Yi, and Zhou]{leng2025stocc}
Ziyang Leng, Jiawei Yang, Wenlong Yi, and Bolei Zhou.
\newblock Occupancy learning with spatiotemporal memory.
\newblock In \emph{Proceedings of the IEEE/CVF International Conference on
  Computer Vision}, pages 26569--26578, 2025.

\bibitem[Li et~al.(2026)Li, Zhang, and Chen]{li2026vqsop}
Feng Li, Chaokun Zhang, and Gong Chen.
\newblock Sparse-aware vector quantization for bandwidth-efficient
  collaborative {3D} semantic occupancy prediction.
\newblock In \emph{Computer Vision -- ECCV 2026}, pages 172--189, 2026.
\newblock Version consulted: arXiv:2607.01928v1.

\bibitem[Li et~al.(2024)Li, He, Zhou, Cheng, Wen, and Zhang]{li2024viewformer}
Jinke Li, Xiao He, Chonghua Zhou, Xiaoqiang Cheng, Yang Wen, and Dan Zhang.
\newblock {ViewFormer}: Exploring spatiotemporal modeling for multi-view {3D}
  occupancy perception via view-guided transformers.
\newblock In \emph{European Conference on Computer Vision}, 2024.

\bibitem[Li et~al.(2023)Li, Yu, Choy, Xiao, Alvarez, Fidler, Feng, and
  Anandkumar]{li2023voxformer}
Yiming Li, Zhiding Yu, Christopher Choy, Chaowei Xiao, Jose~M. Alvarez, Sanja
  Fidler, Chen Feng, and Anima Anandkumar.
\newblock {VoxFormer}: Sparse voxel transformer for camera-based {3D} semantic
  scene completion.
\newblock In \emph{Proceedings of the IEEE/CVF Conference on Computer Vision
  and Pattern Recognition}, pages 9087--9098, 2023.

\bibitem[Li et~al.(2022)Li, Wang, Li, Xie, Sima, Lu, Qiao, and
  Dai]{li2022bevformer}
Zhiqi Li, Wenhai Wang, Hongyang Li, Enze Xie, Chonghao Sima, Tong Lu, Yu Qiao,
  and Jifeng Dai.
\newblock {BEVFormer}: Learning bird's-eye-view representation from
  multi-camera images via spatiotemporal transformers.
\newblock In \emph{European Conference on Computer Vision}, 2022.

\bibitem[Song et~al.(2024)Song, Liang, Cao, Yan, Zimmer, Gross, Festag, and
  Knoll]{song2024cohff}
Rui Song, Chenwei Liang, Hu Cao, Zhiran Yan, Walter Zimmer, Markus Gross,
  Andreas Festag, and Alois Knoll.
\newblock Collaborative semantic occupancy prediction with hybrid feature
  fusion in connected automated vehicles.
\newblock In \emph{Proceedings of the IEEE/CVF Conference on Computer Vision
  and Pattern Recognition}, pages 17996--18006, 2024.

\bibitem[Tang et~al.(2025)Tang, Liu, Sun, Gao, Chen, Xu, and Liu]{tang2025cost}
Zongheng Tang, Yi Liu, Yifan Sun, Yulu Gao, Jinyu Chen, Runsheng Xu, and Si
  Liu.
\newblock {CoST}: Efficient collaborative perception from unified
  spatiotemporal perspective.
\newblock In \emph{Proceedings of the IEEE/CVF International Conference on
  Computer Vision}, pages 1120--1129, 2025.

\bibitem[Wang et~al.(2023{\natexlab{a}})Wang, Zhang, Wang, Zhao, and
  Zhou]{wang2023core}
Binglu Wang, Lei Zhang, Zhaozhong Wang, Yongqiang Zhao, and Tianfei Zhou.
\newblock {CORE}: Cooperative reconstruction for multi-agent perception.
\newblock In \emph{Proceedings of the IEEE/CVF International Conference on
  Computer Vision}, pages 8710--8720, 2023{\natexlab{a}}.

\bibitem[Wang et~al.(2023{\natexlab{b}})Wang, Chen, Chen, Liu, Zhang, Knoll,
  and Jiang]{wang2023umc}
Tianhang Wang, Guang Chen, Kai Chen, Zhengfa Liu, Bo Zhang, Alois Knoll, and
  Changjun Jiang.
\newblock {UMC}: A unified bandwidth-efficient and multi-resolution based
  collaborative perception framework.
\newblock In \emph{Proceedings of the IEEE/CVF International Conference on
  Computer Vision}, pages 8187--8196, 2023{\natexlab{b}}.

\bibitem[Wang et~al.(2024)Wang, Chen, Liao, Fan, and Zhang]{wang2024panoocc}
Yuqi Wang, Yuntao Chen, Xingyu Liao, Lue Fan, and Zhaoxiang Zhang.
\newblock {PanoOcc}: Unified occupancy representation for camera-based {3D}
  panoptic segmentation.
\newblock In \emph{Proceedings of the IEEE/CVF Conference on Computer Vision
  and Pattern Recognition}, pages 17158--17168, 2024.

\bibitem[Wen et~al.(2026)Wen, Huang, He, Meng, and Li]{wen2026mita}
Qishuai Wen, Zhiyuan Huang, Wei He, Xianghan Meng, and Chun-Guang Li.
\newblock Mixture-of-top-$k$ attention: Efficient attention via scalable fast
  weights.
\newblock \emph{arXiv preprint arXiv:2602.01219}, 2026.
\newblock Accepted at NeurIPS 2026; version consulted: v6.

\bibitem[Xiong et~al.(2021)Xiong, Zeng, Chakraborty, Tan, Fung, Li, and
  Singh]{xiong2021nystromformer}
Yunyang Xiong, Zhanpeng Zeng, Rudrasis Chakraborty, Mingxing Tan, Glenn Fung,
  Yin Li, and Vikas Singh.
\newblock {Nystr{\"o}mformer}: A {Nystr{\"o}m}-based algorithm for
  approximating self-attention.
\newblock \emph{Proceedings of the AAAI Conference on Artificial Intelligence},
  35\penalty0 (16):\penalty0 14138--14148, 2021.

\bibitem[Xu et~al.(2022{\natexlab{a}})Xu, Xiang, Tu, Xia, Yang, and
  Ma]{xu2022v2x}
Runsheng Xu, Hao Xiang, Zhengzhong Tu, Xin Xia, Ming-Hsuan Yang, and Jiaqi Ma.
\newblock {V2X-ViT}: Vehicle-to-everything cooperative perception with vision
  transformer.
\newblock In \emph{European Conference on Computer Vision}, 2022{\natexlab{a}}.

\bibitem[Xu et~al.(2022{\natexlab{b}})Xu, Xiang, Xia, Han, Li, and
  Ma]{xu2022opv2v}
Runsheng Xu, Hao Xiang, Xin Xia, Xu Han, Jinlong Li, and Jiaqi Ma.
\newblock {OPV2V}: An open benchmark dataset and fusion pipeline for perception
  with vehicle-to-vehicle communication.
\newblock In \emph{IEEE International Conference on Robotics and Automation},
  2022{\natexlab{b}}.

\bibitem[Xu et~al.(2023)Xu, Tu, Xiang, Shao, Zhou, and Ma]{xu2023cobevt}
Runsheng Xu, Zhengzhong Tu, Hao Xiang, Wei Shao, Bolei Zhou, and Jiaqi Ma.
\newblock {CoBEVT}: Cooperative bird's eye view semantic segmentation with
  sparse transformers.
\newblock In \emph{Proceedings of The 6th Conference on Robot Learning}, pages
  989--1000. PMLR, 2023.

\end{thebibliography}

}
\end{document}